\documentclass[10pt,twocolumn,letterpaper]{article}

\usepackage{cvpr}              

\usepackage[dvipsnames]{xcolor}

\newcommand{\cmark}{\ding{51}}%
\newcommand{\xmark}{\ding{55}}%
\usepackage{arydshln}

\definecolor{golden}{RGB}{255, 215, 0}

\newcommand{\AlgoName}{GRAM }

\newcommand{\Hivt}{Hi-VT5 }
\newcommand{\HivtNoSpace}{Hi-VT5}
\newcommand{\CFormer}{C-Former }
\newcommand{\CFormerNoSpace}{C-Former}
\newcommand{\MPTask}{MP-DocVQA }

\newcommand{\MPData}{MPDocVQA }

\usepackage{graphicx}
\usepackage{amsmath}
\usepackage{amssymb}
\usepackage{booktabs}
\usepackage{multirow}
\usepackage[dvipsnames]{xcolor}
\usepackage{color, colortbl}
\usepackage{pifont}
\usepackage{placeins}
\usepackage[accsupp]{axessibility} 

\definecolor{cvprblue}{rgb}{0.21,0.49,0.74}
\usepackage[pagebackref,breaklinks,colorlinks,citecolor=cvprblue]{hyperref}

\def\paperID{6989} 
\def\confName{CVPR}
\def\confYear{2024}

\title{GRAM: Global Reasoning for Multi-Page VQA}

\author{Tsachi Blau\thanks{Work conducted during an internship at Amazon.}\\
Technion, Israel\\
\and
Sharon Fogel\\
AWS AI Labs\\
\and
Roi Ronen\footnotemark[1]\\
Technion, Israel\\
\and
Alona Golts\thanks{Corresponding author: alongolt@amazon.com}\\
AWS AI Labs\\
\and
Roy Ganz\footnotemark[1]\\
Technion, Israel\\
\and
Elad Ben Avraham\\
AWS AI Labs\\
\and
Aviad Aberdam\\
AWS AI Labs\\
\and
Shahar Tsiper\\
AWS AI Labs\\
\and
Ron Litman\thanks{Corresponding author: litmanr@amazon.com}\\
AWS AI Labs\\
}

\begin{document}
\maketitle

\begin{abstract}
The increasing use of transformer-based large language models brings forward the challenge of processing long sequences. In document visual question answering (DocVQA), leading methods focus on the single-page setting, while documents can span hundreds of pages. We present GRAM, a method that seamlessly extends pre-trained single-page models to the multi-page setting, without requiring computationally-heavy pretraining. To do so, we leverage a single-page encoder for local page-level understanding, and enhance it with document-level designated layers and learnable tokens, facilitating the flow of information across pages for global reasoning. To enforce our model to utilize the newly introduced document tokens, we propose a tailored bias adaptation method. For additional computational savings during decoding, we introduce an optional compression stage using our compression-transformer(\CFormer),reducing the encoded sequence length, thereby allowing a tradeoff between quality and latency. Extensive experiments showcase GRAM's state-of-the-art performance on the benchmarks for multi-page DocVQA, demonstrating the effectiveness of our approach.
\end{abstract}

\section{Introduction}
\label{sec:introduction}

\begin{figure}[ht!]
    \centering
    \includegraphics[width=\linewidth]{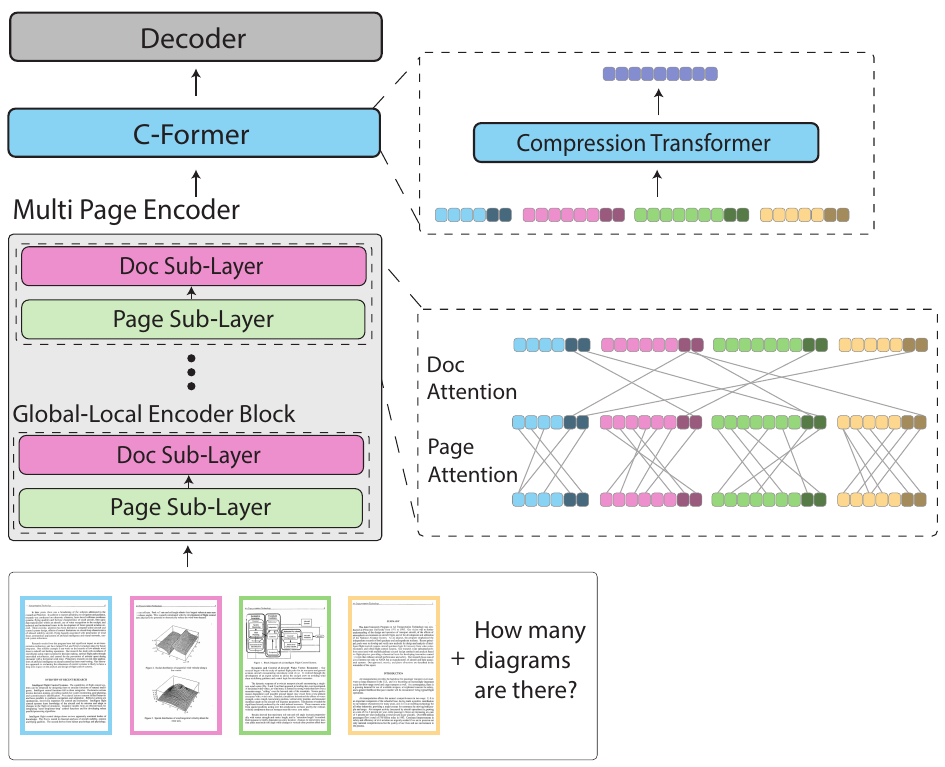}
    \caption{\textbf{An Overview of GRAM.} We suggest an interleaved encoder architecture combining page- with document-attention layers, allowing information to propagate between different pages. An optional compression transformer (C-former) is introduced to allow a trade-off between quality and latency.} 
    \label{Fig:fig_teaser}
\end{figure}

Document understanding, particularly in the context of DocVQA, 
has gained substantial research interest~\cite{xu2020layoutlm, xu2020layoutlmv2, huang2022layoutlmv3, peng2022ernie, appalaraju2021docformer, appalaraju2023docformerv2} and offers a wide array of practical applications, focusing on data extraction and analysis of single page documents.
However, Multi-Page DocVQA (MPDocVQA) poses a more realistic challenge, considering that the majority of documents, including contracts, manuals and scientific papers, often extend well beyond a single page.
Despite the practical relevance of MPDocVQA, it has received limited attention, primarily due to the absence of suitable datasets. Two recently introduced datasets, MPDocVQA~\cite{tito2022hierarchical} and DUDE~\cite{landeghem2023document}, have opened up new avenues for \MPTask research. 

Recent DocVQA approaches rely on transformers~\cite{vaswani2017attention}, at the heart of their architecture. While transformers are a powerful tool, they face challenges when dealing with long input sequences~\cite{press2021train, dai2019transformer, chen2023extending, beltagy2020longformer, zaheer2020big, ainslie2023colt5, bulatov2023scaling}.
This difficulty stems from the self-attention mechanism, which scales quadratically in terms of computation and memory, with respect to the input sequence length.
Addressing this limitation has attracted significant research efforts, primarily in the field of natural language processing (NLP).
Proposed NLP-based solutions can be divided into two main directions: The former aims to modify the attention mechanism to cut computational costs~\cite{beltagy2020longformer, zaheer2020big, ainslie2023colt5}.
The latter involves altering the positional embedding mechanism to improve performance on longer sequences, with minimal fine-tuning~\cite{press2021train, dai2019transformer, chen2023extending}.

A possible option of tackling MPDocVQA is to extend NLP-based approaches to handle multi-modal document data, including visual representations, along with OCR text and corresponding 2D locations and relative page position.
However, this requires extensive pre-training, with relatively scarce multi-page document data, and thus is also sub-optimal in terms of performance. Instead, we opt for leveraging powerful single-page DocVQA models, especially pretrained on millions of single-page documents, and finetuning them to the multi-page scenario. For this purpose, we combine concepts of local (page) and global (document) tokens, which promote an exchange of information within and across pages, while keeping computational cost in check. We choose pages as atomic units in our proposed scheme, as page structure often represent a semantic unit in DocVQA. 

We present GRAM (\textbf{G}lobal \textbf{R}e\textbf{A}soning for \textbf{M}ulti-page VQA), a novel approach for endowing multi-page processing capabilities to existing single-page DocVQA models. Alongside page tokens that encapsulate both textual and visual contents of each page, we introduce doc(ument) learnable tokens, which aim is dispersing global information across all pages. These two sets of tokens interact within our newly-devised two-stage encoder blocks. 
The initial stage utilizes an existing single-page layer and enhances it by including both page and doc tokens as input, allowing them to freely interact.
In the second stage, we prioritize computational efficiency by restricting self-attention solely to the global doc tokens. This global reasoning layer captures collective information from multiple pages, enabling the system to respond to cross-page inquiries, as illustrated in \cref{Fig:fig_teaser}. Considering that doc tokens did not appear in pretraining, to boost their significance during finetune, we employ a designated bias adaptation mechanism which strikes a balance between local and global learnable tokens.

While our method inherently deals with long sequences, we circumvent a quadratic reliance on sequence length by segmenting the document into pages — its semantically logical parts.
We restrict interaction solely among doc learnable tokens, across all pages, thereby mitigating the computational burden of depending quadratically on the page count.
Apart from encoding, the auto-regressive decoding stage poses a computational burden in long sequences. To this end, we introduce a compression stage that precedes the decoder, implemented with a compression transformer, termed \textbf{CFormer}. The CFormer receives the concatenated output of all pages and compresses it to a much shorter sequence, distilling the most pertinent information in the document.
Our key contributions are: 
\begin{itemize}
    \item We propose GRAM, an approach to endow single-page DocVQA methods with multi-page capabilities, without pretraining, allowing the model to process multi-page documents, while preserving single-page performance.  
    \item We introduce document learnable tokens and bias adaptation that enable an effective communication and collaboration between individual pages to support reasoning over multiple page documents.
    \item Our \CFormer module suggests a trade-off between accuracy and compute, distilling information from multi-page sequences into more compact representations.
    \item We obtain SOTA results over the MPDocVQA and DUDE datasets, and provide extensive ablations to each component in our method.
\end{itemize}

\begin{figure*}[ht!]
    \centering
    \includegraphics[width=0.85\textwidth]{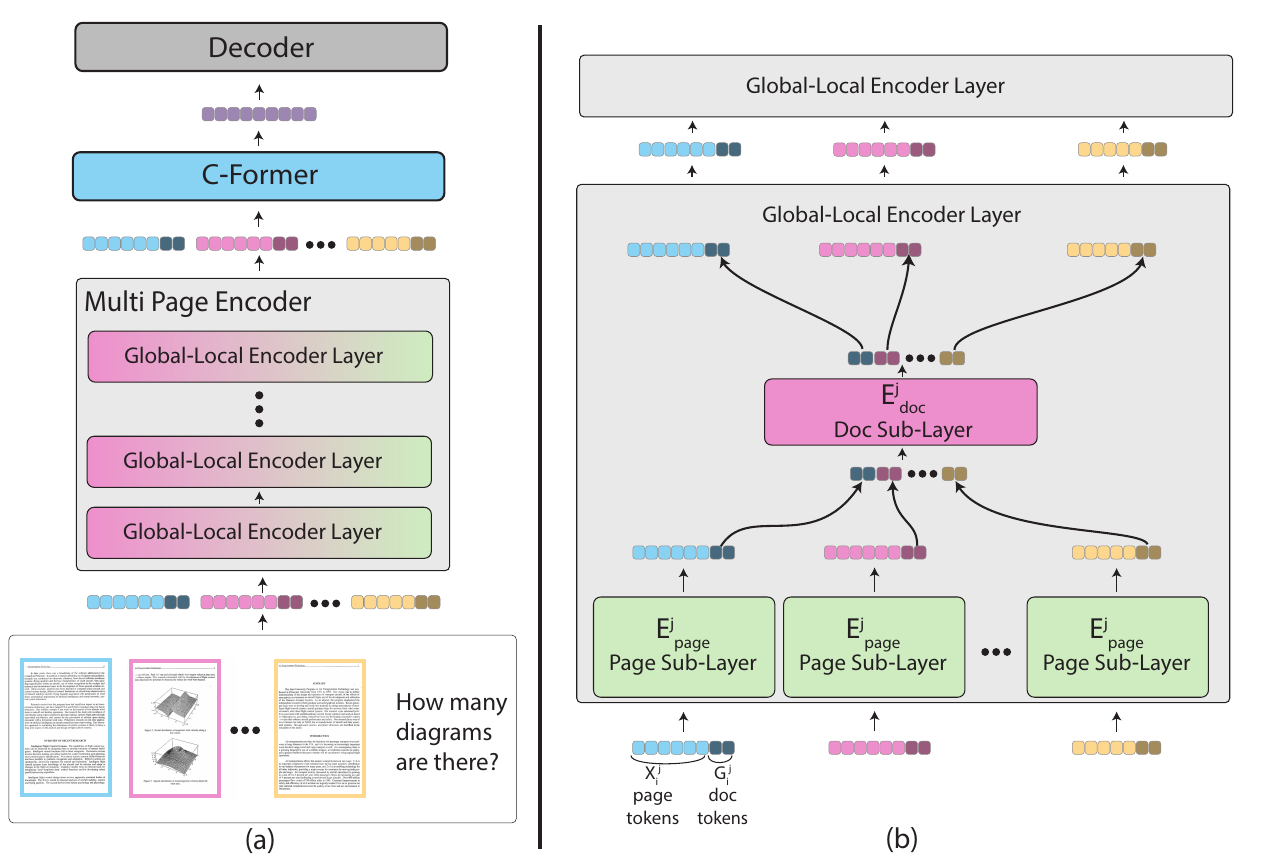}
    \caption{\textbf{GRAM Architecture.} \textbf{(a)} Depicts a high-level architecture overview. For each page, the visual, textual and question tokens are concatenated together with learnable doc tokens (darker color shade). The processed information is fed into the multi-page encoder. The encoder output can be fed directly into the decoder to create the final prediction.  Optionally, a compression model, \CFormer, can be used between the encoder and the decoder to compress the encoder output into a predetermined length, thus reducing overall latency for long documents.
    \textbf{(b)} Shows a global-local encoder layer, containing two sub-layers. The first sub-layer uses self-attention that operates on each page separately, while the second applies a self-attention step on the doc tokens to fuse information between the different pages. The corresponding tokens are then routed back to their respective page and go into the next global-local encoder layer.}
    \label{Fig:fig_arch_overview}
\end{figure*}

\section{Related Work}
\label{sec:background}

\textbf{Long Sequence Approaches} are an active field of research in NLP, aiming to improve the design of chat-systems \cite{openai2023gpt} and image instruction tasks \cite{li2023blip,liu2023visual}. In these applications, the ability to manage and process long sequences is vital, as conversations cannot be cut short, or limited to just a few interactions.
Common approaches to tackle long sequences include sparse attention mechanisms \cite{beltagy2020longformer, zaheer2020big, ainslie2023colt5} and methods to improve results on long sequences during inference \cite{press2021train,raffel2020exploring, chen2023extending}. `Sliding window' approaches of limiting the range of neighbors each token can attend to, lead to a significant reduction in computation and memory consumption. Prominent works of this kind include LongFormer \cite{beltagy2020longformer}, where each token attends to a set of its nearest neighbors, along with additional global tokens. The work of Big-Bird \cite{zaheer2020big} adds additional non-neighboring tokens at random, whereas Colt5 \cite{ainslie2023colt5} uses the same sliding window approach, but performs heavier computations for important tokens and shallow operations for filler words or punctuation. Although Tito \textit{et. al.} \cite{tito2022hierarchical} have demonstrated that the above approaches do not perform as well on the task of MPDocVQA, we do incorporate the ideas of combining both local and global tokens throughout encoding to expand the attention onto additional pages. 

\textbf{DocVQA} has attracted increasing attention \cite{xu2020layoutlm,xu2020layoutlmv2,huang2022layoutlmv3,appalaraju2021docformer,appalaraju2023docformerv2,tang2023unifying,powalski2021going,peng2022ernie,biten2022latr} with the introduction of the DocVQA dataset by Mathew et al. \cite{mathew2021docvqa}. Most methods in DocVQA leverage OCR~\cite{litman2020scatter, aberdam2021sequence,nuriel2022textadain,aberdam2022multimodal,aberdam2023clipter} to input both text and layout information (bounding box coordinates and possibly font type) into the model, where some further explore different techniques to combine the two types of data streams, or alternatively, clever schemes of pretraining. DocVQA methods can be roughly divided to two categories: extractive and abstractive. Extractive methods \cite{peng2022ernie,xu2020layoutlm,xu2020layoutlmv2,huang2022layoutlmv3} rely on the fact that the explicit answer resides in the written text, thus only output a corresponding text span within the input sequence. Abstractive methods \cite{tang2023unifying, powalski2021going,appalaraju2021docformer,appalaraju2023docformerv2,ganz2023towards}, on the other hand, have the capacity to generate free-form answers which do not necessarily appear in the text, thus providing flexibility in real-world applications. Notably, existing research in DocVQA does not scale in a straightforward way to deal with the more realistic multi-page scenario.

\textbf{MPDocVQA} has recently gained momentum with the launch of two new multi-page datasets: MP-DocVQA \cite{tito2022hierarchical} and DUDE \cite{landeghem2023document}, offering two separate recipes to tackle longer documents. The first approach of Tito \textit{et. al.}, referred to as HiVT5 \cite{tito2022hierarchical}, suggests compressing the encoding of each page separately, and feeding the decoder with the concatenation of the compressed outputs from each single-page encoder. While this approach is advantageous in terms of computation, we later show the compression may severely hinder the results. In addition, there is no communication between the single-page encoders until the final stage of decoding, whereas in our method, we allow for exchange of page and document information throughout all stages of encoding. Another prominent approach, proposed by Landeghem \textit{et. al.} \cite{landeghem2023document}, which relatively preserves quality, involves concatenating all the pages into one long sequence and feeding it to a standard encoder-decoder structure. This, however, poses a heavy computational burden as transformers' self-attention component scales quadratically with input sequence length.

\section{GRAM}
\label{sec:our_method}

\subsection{Base Architecture}
\label{arch}

The underlying idea in our approach is using existing encoder-decoder single-page models for document understanding and extending them to multi-page scenarios, without additional pretraining. In this work, we provide such a recipe over the notable DocFormerv2 \cite{appalaraju2023docformerv2}. For the sake of completeness, DocFormerv2 is a T5-based \cite{raffel2020exploring} encoder-decoder transformer model which operates over both visual and textual features to support document understanding. Each page is represented by textual features $\textbf{T} \in R^{N_t \times d}$ which encapsulate OCR tokens and their corresponding 2D bounding box positions \cite{xu2020layoutlm}, along with visual $\textbf{V} \in R^{N_v \times d}$ and question $\mathbf{Q} \in R^{N_q \times d}$ embeddings. Where $N_t$, $N_v$ and $N_q$ are the lengths of the OCR, visual features and the question. The output result $\textbf{Y}$ is obtained by passing the concatenated inputs through the encoder-decoder model,
\begin{equation}
    \textbf{Y} = \textbf{D}(\textbf{E}(\operatorname{concat}( \textbf{T}, {\textbf{V}, \textbf{Q}}))),
\end{equation}
where \textbf{E} and \textbf{D}, are the encoder and decoder, respectively. 

Our method uses these basic building blocks in designing a multi-page solution. To this end, we introduce a bi-level global-local encoder, as illustrated in \cref{Fig:fig_arch_overview}. At the local page-level of each block, we utilize the layers of the existing single-page encoder $\textbf{E}$ to process each page separately, together with learnable doc tokens. Next, we introduce a slim global layer in each block that facilitates communication between doc tokens across all pages. 
This bi-level localized processing ensures the model can understand the content of each page effectively, while also combining information across pages in the document. After $M$ such blocks, we feed the encoded features from all pages into the existing decoder $\textbf{D}$ to produce the overall output.

\subsection{Global-Local Reasoning}
\label{global}

To operate on multiple pages we break down the document to $K$ pages, and the single-page encoder to $M$ encoder layers, $\textbf{E}^j, j=0,...,M-1$. We then construct $M$ blocks, with two sub-layers each. The first page sub-layer originates from the existing pretrained encoder layer, referred to as $\textbf{E}^j_{page}$, and operates in parallel, with shared weights, for all pages in the document. This layer receives both page and doc tokens. The second, newly introduced, document layer $\textbf{E}^j_{doc}$ collects only the doc tokens from all pages and promotes sharing information across all of the document. 

Formally, we augment the input of the standard single-page encoder with page-specific indexing $(\textbf{T}_i,\textbf{V}_i,\textbf{Q})$ and incorporate page-positional embedding $\textbf{P}_i$ to both text and visual features, where $i=0,...,K-1$:
\begin{equation}
\tilde{\mathbf{T}}_i = \mathbf{T}_i + \mathbf{P}_i, \quad \tilde{\mathbf{V}}_i = \mathbf{V}_i + \mathbf{P}_i,.
\end{equation}

Next, we formulate our bi-level global-local block. The input to the first page-level sub-layer in each
block is the concatenation of the textual, visual and question features, denoted $\mathbf{X}_i^j = \operatorname{concat}(\tilde{\mathbf{T}}_i, \tilde{\mathbf{V}}_i, \mathbf{Q}_i)$, along with page-specific doc tokens $\mathbf{G}^j_i \in R^{N_g \times d}$,

\begin{equation} \label{eq:stage_one}
    \textbf{X}_i^{j+1}, \tilde{\mathbf{G}}_i^{j+1} = \mathbf{E}^j_{page}(\operatorname{concat}(\textbf{X}_i^j, \textbf{G}_i^j)).
\end{equation}

Here, the features undergo self-attention, normalization and feed-forward layers. 
The layer output $\mathbf{X}_i^{j+1}$ is passed on as input to the next bi-level block, whereas only the doc tokens $\tilde{\mathbf{G}}_i^{j+1}$, enter the second doc sub-layer, which again includes self-attention, normalization and feed-forward
\begin{equation} \label{eq:stage_two}
    \{\mathbf{G}_i^{j+1}\}_{i=0}^{K-1} = \mathbf{E}^j_{doc}(\operatorname{concat}(\{\tilde{\mathbf{G}}_i^{j+1}\}_{i=0}^{K-1})).
\end{equation}

In this stage, the doc tokens can interact and pass information from page to page, after which being passed on to the next block, as depicted in \cref{eq:stage_one}. This design allows information to flow between pages while keeping computational costs in check.
When concluding the traversal over $M$ such layers, the outputs across all pages are concatenated and fed to the decoder $\textbf{D}$,
\begin{equation}
    \mathbf{Y}_{multi} = \mathbf{D} (\operatorname{concat}(\{\mathbf{X}_i^{M}, \mathbf{G}_i^{M}\}_{i=0}^{K-1})).
\end{equation}

\begin{figure}[t]
    \centering
    \includegraphics[width=0.9\linewidth]{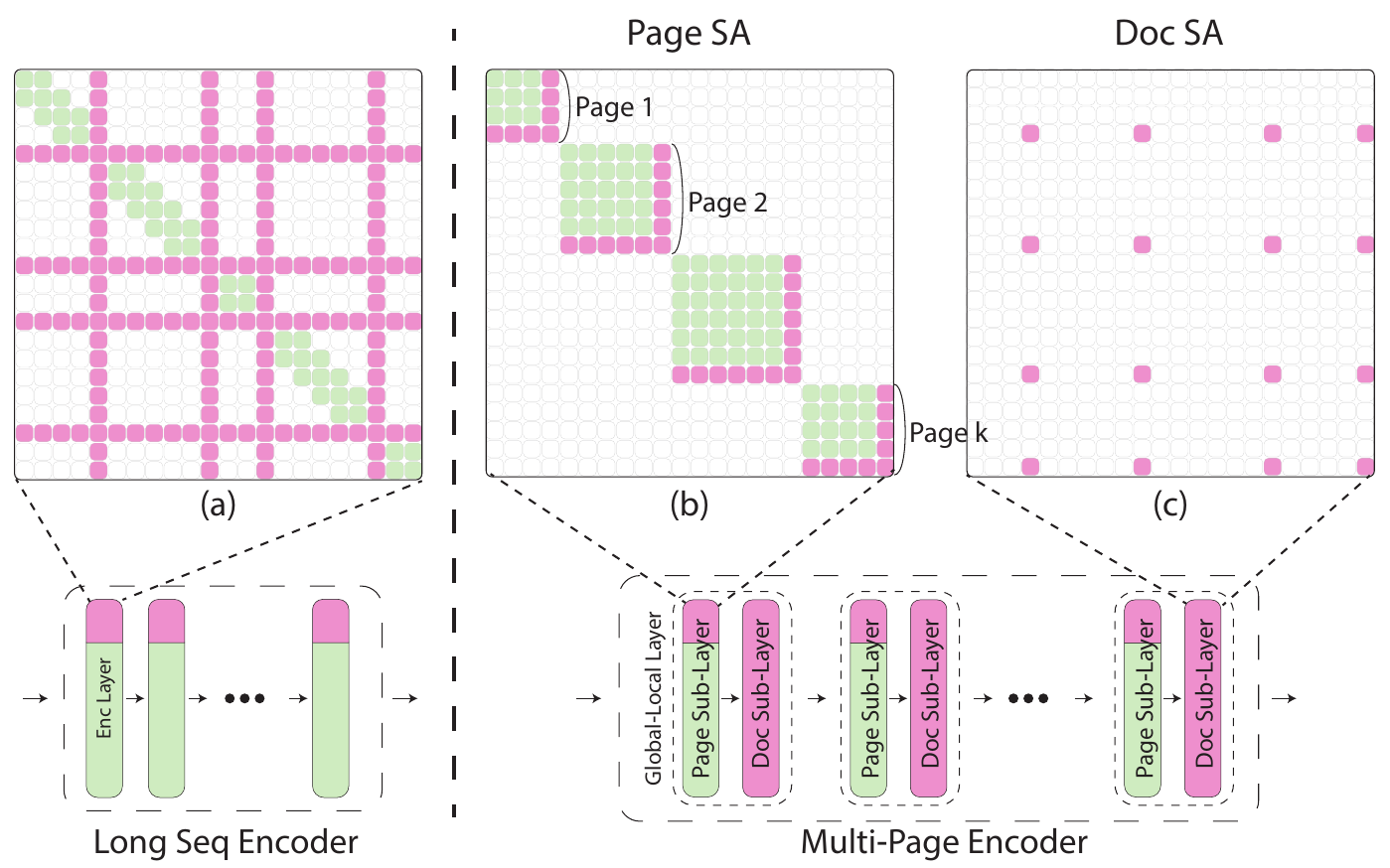}
    \caption{
    \textbf{Global-Local Attention}: In long sequence approaches (a), attention is applied jointly to the entire sequence of concatenated local and global tokens. Our method, separates the computation into two steps — page-level (b) and document-level (c)— leveraging the natural division of documents into pages.}
    \label{Fig:fig_attention_mask}
    
\end{figure}

To visualize the difference between the attention masks in our method, we compare it with previous long sequence approaches~\cite{beltagy2020longformer, zaheer2020big, ainslie2023colt5} in \cref{Fig:fig_attention_mask}. These prior methods optimize computation by using attention masking on nearby tokens and allowing limited global connections. However, naively applying such methods to multi-page documents will treat it as a single stream, which does not consider the division into pages. Our global-local blocks, with a two-stage attention-masking mechanism, better suit multi-page documents. In addition, our two-level design benefits from existing, extensively pretrained single page models.

\subsection{Bias Adaptation}
\label{bias}

An already-pretrained model, introduced with a new stream of data, might disregard it altogether~\cite{french1999catastrophic,tsimpoukelli2021multimodal,zhai2022lit,ganz2023towards}.
To overcome this, we force the system to account for the newly-introduced doc tokens by modifying the encoder's bias method.
Originally, the bias method intervenes in the attention mechanism, diminishing the relationships between distant tokens.
However, in our specific case, the distance between doc and page tokens does not represent their actual relevance.
To enforce the encoder to pay closer attention to the doc tokens, we assign them a positive constant bias value.
Particularly, we replace the values in the bias matrix, corresponding with the doc tokens, with fixed ones.
Instead of a single bias value, we utilize a different value for each attention head, as performed in ALiBi~\cite{press2021train}, enabling more fine-grained control of the global features in each head.
Specifically, the constant doc bias value is set to $c \cdot \frac{1}{2^a}$, where $c$ is a constant and $a$ is the attention head index.
This yields a decaying bias value across different attention heads, resulting in hierarchical importance of the document  information, where the first heads are more oriented towards doc tokens and the last towards page tokens.

\subsection{Compression Transformer}
\label{compression}
Our global-local solution to \MPTask resolves the problematic quadratic dependency on the number of pages $K$ during encoding.
However, the auto-regressive decoding complexity scaling linearly with $K$ also poses a practical challenge during inference time, as we later discuss in \cref{compute_analysis}.
To alleviate this burden, we place an optional transformer-based model, named \CFormer (\textbf{C}ompression Trans\textbf{Former}), between the encoder outputs and the decoder, as depicted in \cref{Fig:fig_arch_overview}. The \CFormer has the ability to revise the information across all pages and distill only the important details, required to correctly answer the question. 

Specifically, the \CFormer is a light-weight transformer-based decoder \cite{raffel2020exploring}, denoted as $\mathbf{D}_C$, featuring cross-attention, layer norm and feed-forward layers in each block.
The input to \CFormer includes $N_c$ learnable tokens $\mathbf{C} \in R^{N_c \times d}$, concatenated with the input question $\tilde{\textbf{C}}=\operatorname{concat}( \textbf{C}, \textbf{Q})$. 
In addition, we feed it with the outputs of the global-local interlaced encoder, concatenated to one long sequence, referred as $\mathbf{O}$, where $\mathbf{O} = \operatorname{concat}(\{\mathbf{X}_i^{M}, \mathbf{G}_i^{M}\}_{i=0}^{K-1})$.
The output of \CFormer is thus
\begin{equation*}
        \textbf{O}_C = \textbf{D}_C(Q{=}\tilde{\textbf{C}}, K{=}\mathbf{O}, V{=}\mathbf{O}),
\end{equation*}
where we pass forward only the first set of $N_c$ output embeddings and ignore the rest, setting the output sequence dimension to $N_c$.
\CFormer offers flexibility in controlling the tradeoff between ANLS quality and computational efficiency by controlling the output sequence length $N_c$.

\subsection{Computation Analysis}
\label{compute_analysis}
Next, we turn to provide a thorough computational complexity analysis.
We consider a document that comprises of $K$ pages, each with $N$ tokens, and the maximum answer length is $L$. For simplicity, we assume that all encoders and decoders have one layer. The naïve way to support multi-page documents is using an existing single-page encoder-decoder model, fusing all of the textual page inputs together, and feeding them as one long sequence. We refer to this approach as `concat'. The self-attention complexity of such a configuration scales quadratically with the sequence length, $O((N \cdot K)^2)$. Conversely in our method, we operate on the document pages with two alternating encoding stages in each layer. The first stage performs a self-attention over both the page and doc tokens.
Hence, the complexity of such sub-layer is $O((N + N_g)^2 \cdot K)$, where $N_g$ is the number of doc tokens. The second stage features a self-attention operation over the doc tokens, across all pages in the document. The complexity of this operation is {$O((N_g \cdot K)^2)$}. Overall, the total complexity for one global-local encoder block is $O((N + N_g)^2 \cdot K + (N_g \cdot K)^2)$. Since $N_g$ is a constant, and the number of pages is usually less than the number of words in each page ($K<N$), 
we obtain a complexity of $O(N^2 \cdot K)$, which is not quadratic in $K$.

Prior to decoding, the outputs of all per-page encoders are concatenated, thus the output sequence length is ${(N+N_g) \cdot K}$. Since the decoder is auto-regressive, its complexity depends quadratically on the maximum output length, $L$, namely, ${O((N+N_g) \cdot K \cdot L^2) = O(N \cdot K \cdot L^2)}$. Since this operation of decoding is performed iteratively during inference, the combined sequence length $(N+N_g) \cdot K$ becomes computationally heavy. To alleviate this concern, we propose an optional \CFormer model, which performs compression prior to decoding. The overall complexity in this decoding scheme includes passing through the \CFormer and then through the decoder, leading to $O((N+N_g) \cdot K \cdot N_c) + N_c \cdot L^2)$ which is equivalent to $O(N \cdot K + L^2)$, since $N_c$ is a constant, denoting the number of compression tokens in \CFormerNoSpace.

\begin{table*}[ht]
\small
  \centering
  \bgroup
  \def\arraystretch{1}
  \resizebox{1\linewidth}{!}{%
  \begin{tabular}{l c | c | c c c c c}
    \toprule
    \multirow{3}{*}{\textbf{Method}} & \multirow{3}{*}{\textbf{Params}} & \multicolumn{1}{c}{\textbf{MPDocVQA}} & \multicolumn{5}{c}{\textbf{DUDE}} \\
    & & \multirow{2}{*}{\textbf{ANLS}}  & \multirow{2}{*}{\textbf{ANLS}} & \multicolumn{4}{c}{\textbf{ANLS per Question Type}} \\
    & & & & Extractive & Abstractive & List of answers & Unanswerable \\
    \hline
    Longformer \cite{beltagy2020longformer} & $148M$ & $55.06$ & $27.14$ & $43.58$ & $8.55$ & $10.62$ & $10.78$ \\
    BigBird \cite{zaheer2020big} & $131M$ & $58.54$ & $26.27$ & $40.26$ & $7.11$ & $8.46$ & $12.75$\\
    LayoutLMv3 \cite{zaheer2020big} & $125M$ & $55.13$ & $20.31$ & $32.60$ & $8.10$ & $7.82$ & $8.82$ \\

    \HivtNoSpace $_{beamsearch}^\dagger$ \cite{HIVT5beam} & $316M$ & $-$ & $35.74$ & $28.31$ & $32.98$ & $10.60$ & $62.90$\\
    \HivtNoSpace \cite{tito2022hierarchical} & $316M$ & $62.01$ & $23.06$ & $17.60$ & $33.94$ & $6.83$ & $61.67$ \\
    \HivtNoSpace* & $257M$ & $60.78$ & $23.86$ & $7.21$ & $16.56$ & $3.53$ & $72.77$ \\
    $\text{DocFormerv2}_{concat}$ \cite{appalaraju2023docformerv2} & $257M$ & $69.67$ & $44.21$ & $41.66$ & $41.86$ & $15.13$ & $\mathbf{65.19}$ \\
    \rowcolor{golden!20} $\text{GRAM}_{C-Former}$ & $286M$ & $70.80$ & $40.07$ & $40.43$ & $39.61$ & $11.42$ & $52.55$ \\
    \rowcolor{golden!20} GRAM & $281M$ & $\mathbf{73.68}$ & $\mathbf{46.15}$ & $\mathbf{46.07}$ & $\mathbf{44.82}$ & $\mathbf{15.27}$ & $62.18$ \\

    \hline
    T5-2D \cite{landeghem2023document} & $770M$ & $-$ & $46.06$ & $\mathbf{55.65}$ & $\mathbf{50.81}$ & $5.43$ & $\mathbf{68.62}$  \\
    DocGptVQA \cite{DocGptVQA} & $>3.5B$ & $-$ & $50.02$ & $51.86$ & $48.32$ & $28.22$ & $62.04$  \\
    DocBlipVQA \cite{DocBlipVQA} & $>3.5B$ & $-$ & $47.62$ & $50.69$ & $46.31$ & $\mathbf{30.73}$ & $55.22$\\
    Hi-VT5* \cite{tito2022hierarchical} & $784M$ & $71.35$ & $28.89$ & $18.21$ & $26.17$ & $6.84$ & $58.99$  \\
    $\text{DocFormerv2}_{concat}$ \cite{appalaraju2023docformerv2} & $784M$ & $76.40$ & $48.44$ & $50.82$ & $48.06$ & $17.67$ & $59.04$  \\
    \rowcolor{golden!20} $\text{GRAM}_{C-Former}$  & $864M$ & $77.60$ & $45.47$ & $47.63$ & $44.91$ & $14.34$ & $56.99$  \\ 	
    \rowcolor{golden!20} GRAM & $859M$ & $\mathbf{80.32}$ & $\mathbf{51.15}$ & $53.67$ & $50.35$ & $18.40$ & $63.23$  \\

    \hline
    \HivtNoSpace*$^\dagger$ \cite{tito2022hierarchical} & $784M$ & $73.51$ & $49.18$ & $49.29$ & $48.35$ & $13.30$ & $\mathbf{65.95}$  \\
    $\text{DocFormerv2}_{concat}^\dagger$ \cite{appalaraju2023docformerv2} & $784M$ & $76.77$  & $50.79$ & $52.70$ & $49.61$ & $17.33$ & $65.14$ \\
    \rowcolor{golden!20} $\text{GRAM}_{C-Former}^\dagger$  & $864M$ & $78.12$ & $50.97$ & $55.15$ & $50.46$ & $17.26$ & $61.04$  \\ 	
    \rowcolor{golden!20} GRAM$^\dagger$ & $859M$ & $\mathbf{79.67}$ & $\mathbf{53.36}$ & $\mathbf{56.83}$ & $\mathbf{52.32}$ & $\mathbf{19.96}$ & $65.43$ \\

    \bottomrule
    
  \end{tabular}
  }
  \egroup
  \caption{\textbf{Quantitative Results}. We present ANLS results for the \MPData \cite{tito2022hierarchical} and DUDE \cite{landeghem2023document} test sets. The methods are grouped according to the model type and size, starting from encoder-only models (top), T5-base models (middle) and T5-large models (bottom). $^\dagger$ denotes training with both \MPData and DUDE.}
  \label{table:MPDocVQA_DUDE}
\end{table*}

\section{Experiments}
\label{sec:experiments}
\tabcolsep=0.11cm
\begin{table}[]
  \centering
  \bgroup
  \def\arraystretch{1.1}
  \resizebox{1\linewidth}{!}{

  \begin{tabular}{l | c  c | c c}
    \toprule
    \multirow{2}{*}{\textbf{Method}} & \multicolumn{2}{c}{\textbf{Training Data}} & \multicolumn{2}{c}{\textbf{ANLS}} \\
        & \textbf{DocVQA} & \textbf{MPDocVQA} & \textbf{DocVQA} & \textbf{MPDocVQA} \\
    \hline
    
     \multirow{3}{*}{\textbf{$\text{DocFormerv2}_{concat}$}}  & \cmark & \xmark & $86.60$ & $72.73$ \\
      & \xmark & \cmark & $85.28$ & $76.40$ \\
      & \cmark & \cmark & $86.47$ & $75.37$ \\
     \hline
    \multirow{3}{*}{\textbf{GRAM}} & \cmark & \xmark & $86.70$ & $73.12$ \\
     & \xmark & \cmark & $85.29$ & $80.32$ \\
     & \cmark & \cmark & $86.32$ & $78.66$ \\

    \bottomrule
  \end{tabular}
  }
  \egroup
  \caption{\textbf{DocVQA vs. MPDocVQA Performance.} Test results over both datasets using the large model variants. A checkmark denotes whether a dataset was included in training or not.}
  \label{table:DocVQA}
\end{table}


\subsection{Experimental Settings}

\paragraph{Datasets and Metrics}
The \MPData dataset \cite{tito2022hierarchical} features 46K questions, spanning over 48K images, and includes layout elements as figures, tables, lists and diagrams, with printed, handwritten and typewritten text. MPDocVQA contains mostly extractive questions, for which answers are present in the given text. DUDE is smaller in size (23.7K questions over 3K documents), but offers complex questions that require a reader to rationalize beyond the written text content.
We report our results using the ANLS metric, introduced in \cite{biten2019scene}, computing a generalized accuracy. 
Results for DUDE can be broken apart to several types of questions, categorized to four groups: `extractive' -- for which the answer is found directly in the text; `abstractive' -- requiring a free-form answer that does not necessarily appear in the document; `list of answers' -- requiring a list of answers, as opposed to a single one, and `unanswerable' -- where the result cannot be determined using the text.

\paragraph{Implementation Details}

Our underlying architecture is based on Docformerv2 \cite{appalaraju2023docformerv2}. Recall, our interlaced encoder features $M$ blocks ($12$ in `base' and $24$ in `large'), where each block contains a page sub-layer which originates from an extension of Docformerv2's encoder layer. Every structure contains self-attention, normalization and feed-forward layers. We extend the page layer from Docformerv2 to feature also the doc learnable embeddings. The second doc sub-layer is similar in structure to the first sub-layer, only it is initialized from scratch, with the following specification: $d_{ff}=1024$, $d_{kv}=64$, $nheads=4$, $d=256$. We implement $32$ doc learnable tokens for each page, uniformly initialized to random values. For bias adaptation, the initial bias value is set to $c=20$, with variations between encoder heads, as described in \cref{bias}. We incorporate an additional optional compression stage using \CFormer-- a randomly-initialized T5 \cite{raffel2020exploring} tiny decoder, with an encoder mask instead of a causal one. The output sequence extracted from \CFormer is $N_c=256$.
Finally the decoder is initialized with pretrained weights from DocformerV2.

The model is trained with the Hugging Face Trainer \cite{HuggingFaceTrainer} for $200k$ steps, starting with a warm-up of $1k$ steps, with linear learning rate decay. We use learning rates of $3e^{-5}$ and $1e^{-4}$ for the already pretrained encoder and decoder weights, versus the newly initialized doc sub-layer weights.
Training is performed on a cluster of $8\times A100$ GPUs, each with $40GB$ of RAM. During training, each page encoder receives $800$ tokens, dealing with up to $4$ pages. During testing, we increase the maximum length of tokens to $8,000$. 

\begin{figure*}[t]
    \centering
    \includegraphics[width=0.9\textwidth]{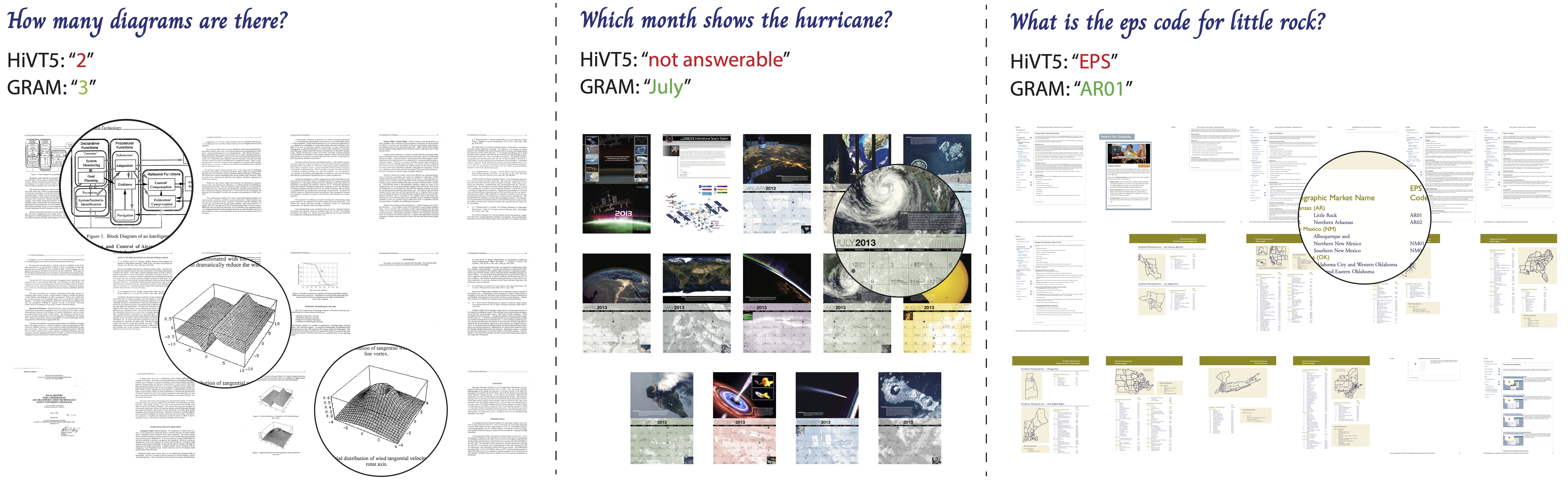}
    \caption{Qualitative comparison between our approach and \Hivt \cite{tito2022hierarchical} indicate that the integration of our global-local encoder enhances reasoning capabilities, especially when the inquiries require multi-page context.}
    \label{Fig:data_examples}
\end{figure*}

\paragraph{Baselines} 
We report the results of previous work on both MP-DocVQA and DUDE datasets (if those exist), including the NLP-based Longformer \cite{beltagy2020longformer} and BigBird \cite{zaheer2020big}, which were adapted to MPDocVQA by \cite{tito2022hierarchical}; LayoutLMv3 \cite{huang2022layoutlmv3}, originally designed for DocVQA; and Hi-VT5 \cite{tito2022hierarchical} and T5-2D \cite{tito2021document}, specifically suggested for \MPTask Task. We also add for reference the results of methods published in the leader-boards of \MPData and DUDE, which do not have corresponding papers, including DocGptVQA \cite{DocGptVQA}, DocBlipVQA \cite{DocGptVQA}, and $\text{Hi-VT5}_{beamsearch}$ \cite{HIVT5beam} (\cite{HIVT5beam} was trained on both MP-DocVQA and DUDE). 
In our approach, we present two variations: GRAM and GRAM$_{C-Former}$. While GRAM utilizes the full length of the encoder output, GRAM$_{C-Former}$ allows the user to control the trade-off between performance and latency.

To ensure  a fair comparison, since we use the pretrained model of DocFormerv2 \cite{appalaraju2023docformerv2}, we implement two additional baselines, referred to as \HivtNoSpace* and $\text{DocFormerv2}_{concat}$. The first follows a similar structure as \Hivt \cite{tito2022hierarchical}, with the encoder originating from DocFormerv2, however without the page answer prediction, as it does not exist in DUDE. The second recreates the approach of \cite{landeghem2023document}, where only the textual tokens of all pages are concatenated to one long sequence, then passing through the DocFormerv2 model. The second approach poses a computational burden, thus we use only $600$ tokens during training per page, with up to $4$ pages, and during test only $400$ tokens.

\subsection{Results}
We present the performance of our method over the \MPData \cite{tito2022hierarchical} and DUDE \cite{landeghem2023document} datasets in \cref{table:MPDocVQA_DUDE}. The methods are divided into three groups: the top contains encoder-only, and methods that rely on the T5-base model (up to $316M$ parameters); the middle section, approaches that use the T5-large model (over $770M$ parameters), and finally the bottom, T5-large models, trained on both datasets. 

As can be seen, in the first group, the encoder-only NLP methods, LongFormer \cite{beltagy2020longformer}, BigBird \cite{zaheer2020big} and LayoutLMv3 \cite{huang2022layoutlmv3} can only handle relatively well `extractive' style tasks as in \MPData dataset \cite{tito2022hierarchical}, but often struggle with `abstractive' questions that are more abundant in DUDE \cite{landeghem2023document}. As to T5-`base' models, versus our best competitor Docformerv2$_{concat}$, we obtain an improvement of $(+4\%, +1.9\%)$ on MP-DocVQA and DUDE datasets. As to methods that combine an additional compression before decoding (Hi-VT5, Hi-VT5*), our \CFormer achieves an increase in $(+8.8\%, +16.2\%)$ over the best candidates on the MP-DocVQA and DUDE datasets.

As for the group of `large' models, we include the results of T5-2D \cite{landeghem2023document} DocGptVQA \cite{DocGptVQA} and DocBlipVQA \cite{DocBlipVQA}. Note that our model surpasses DocFormer$_{concat}$, the primary baseline, achieving improvements of $(+3.9\%,+2.7\%)$ on MP-DocVQA and DUDE, respectively. We also outperform DocGptVQA \cite{DocGptVQA}, a method that appears in the leaderboard of DUDE, by $+1.1\%$, thereby obtaining SOTA results for GRAM `large'.

The final category showcases large encoder-decoder models, fine-tuned on both MP-DocVQA and DUDE training sets, showcasing the benefits of augmented training data. \AlgoName consistently demonstrates performance gains over the baseline, illustrating its robustness across different datasets and training scenarios. Next, we present in \cref{table:DocVQA} the effect of training on DocVQA vs. \MPData. Our method achieves performance on-par on the single page task, while enhancing performance on the multi-page scenario by $+3.3\%$, compared to the baseline.

In \cref{Fig:data_examples}, We show qualitative results on the DUDE dataset of \AlgoName versus Hi-VT5* \cite{tito2022hierarchical}. Our method demonstrates proficiency in addressing questions that involve attention over multiple pages (\emph{`how many diagrams are there'}), an increased visual analysis capability (\emph{`Which month shows the hurricane?'}), and heightened abstractive ability (\emph{`What is the EPS code for Little Rock?'}).

\section{Ablation Study}
\label{sec:ablation}

\tabcolsep=0.09cm
\begin{table}[]
  \centering
  \bgroup
  \def\arraystretch{1.1}
  \resizebox{1\linewidth}{!}{
  \begin{tabular}{c c c | c c c c c}
    \toprule
        \multirow{3}{*}{\textbf{\shortstack{\#Doc\\ Tokens}}} & \multirow{3}{*}{\textbf{\shortstack{Bias \\Type}}} & \multirow{3}{*}{\textbf{\shortstack{Compression \\Dimension}}} & \multicolumn{5}{c}{\textbf{ANLS by Number of Pages}} \\
        & & & \multicolumn{5}{c}{\textbf{DUDE validation dataset}} \\
        & & & All & 1 & 2-4 & 5-10 & 11-end \\
            \hline
            \rowcolor{lightgray!20}  \xmark & \xmark & \xmark & $46.16$ & $47.18$ & $48.66$ & $43.34$ & $42.57$ \\
            
            \hline
            
            16 &  & \xmark & $46.39$ & $48.35$ & $49.06$ & $43.16$ & $41.56$ \\
            \rowcolor{golden!20} 32 & \multirow{1}{*}{Decaying} & \xmark & $\mathbf{47.88}$ & $\mathbf{49.29}$ & $\mathbf{49.90}$ & $\mathbf{45.90}$ & $\mathbf{43.94}$ \\
            64 &  & \xmark & $46.70$ & $47.98$ & $49.22$ & $44.00$ & $42.60$ \\
            
            \hline
            \multirow{3}{*}{32} & \xmark & \xmark & $47.52$ & $\mathbf{49.85}$ & $\mathbf{49.93}$ & $44.90$ & $42.10$ \\
             & Constant & \xmark & $46.14$ & $47.41$ & $48.13$ & $44.44$ & $42.19$ \\            
            \rowcolor{golden!20}  & Decaying & \xmark & $\mathbf{47.88}$ & $49.29$ & $49.90$ & $\mathbf{45.90}$ & $\mathbf{43.94}$ \\

            \hline
            
            \multirow{5}{*}{32} & \multirow{5}{*}{Decaying} & 8 & $39.83$ & $39.73$ & $41.41$ & $36.98$ & $39.52$ \\
             &  & 32 & $40.42$ & $40.95$ & $41.64$ & $38.12$ & $39.39$ \\
             &  & 256 & $41.99$ & $42.57$ & $43.77$ & $38.40$ & $41.01$ \\
             &  & 1024 & $42.56$ & $42.97$ & $44.30$ & $38.94$ & $41.93$ \\    
             &  & 4096 & $\mathbf{43.59}$ & $\mathbf{44.75}$ & $\mathbf{44.64}$ & $\mathbf{40.54}$ & $\mathbf{42.67}$ \\


    \bottomrule
  \end{tabular}
  }
  \egroup
  \caption{\textbf{\AlgoName Ablation Study}. Results on DUDE validation set ablating over (a) the dimension of doc tokens, (b) the attention bias employed and (c) the C-former input dimension.}
  \label{table:DUDE_ablation}
\end{table}

We perform an ablation study on our approach, evaluating the influence of each constituent component using DUDE's validation set \cite{landeghem2023document}. This validation set enables the grouping of documents by their respective page counts: 1, 2--4, 5--10, 11--end, encompassing $1747, 2259, 1062, 1241$ samples in each category, respectively. Our investigation delves into the impact of the number of doc tokens and the bias adaptation methods. Moreover, we employ the \CFormer for sequence compression, adjusting the compression ratio and examining the balance between performance and latency (see supplementary for more details). 

\paragraph{\AlgoName Components} We focus our initial exploration on the impact of the number of doc tokens $N_g$. As can be seen in \cref{table:DUDE_ablation}, while $N_g=16$ leads to performance on-par with not using doc tokens at all, for the optimal value of $N_g=32$, we obtain an increase of $+1.7\%$ in ANLS.
Shifting our focus to bias adaptation methods, \cref{table:DUDE_ablation} shows that using constant bias has a negative effect on the results, suggesting this method is not flexible enough in maintaining a balance between the page and doc tokens. However, our decaying bias-adaptation approach does improve results overall, versus no-bias $(+0.36\%)$, especially for longer documents ($+1\%$ improvement for 5-10 pages and $+1.84\%$ for 11 pages and more). This is to be expected, since incorporating new doc tokens and increasing their importance can potentially affect single-page performance.
Finally, in \cref{table:page_ablation}, we reinforce our choice of pages as semantic logical units for MPDocVQA. We first ablate our method with and without page embedding. Next, we compare our page-based division with varying fixed-length division of tokens for encoder. Results in \cref{table:page_ablation} clearly demonstrate an advantage towards page-level encoding in MPDocVQA. This aligns with our initial assumption that structured documents are often designed with page-division in mind.

\tabcolsep=0.09cm
\begin{table}[]
  \centering
  \bgroup
  \def\arraystretch{1.0}
  \resizebox{1\linewidth}{!}{
  \begin{tabular}{c c | c c c c c}
    \toprule
        \multirow{3}{*}{\textbf{\shortstack{Page\\ Embedding}}} & 
        \multirow{3}{*}{\textbf{\shortstack{Segment \\Length}}} &  
        \multicolumn{5}{c}{\textbf{ANLS by Number of Pages}} \\
        & 
        & 
        \multicolumn{5}{c}{\textbf{DUDE validation dataset}} \\
        & & All & 1 & 2-4 & 5-10 & 11-end \\
            \hline
            \rowcolor{golden!20}
            \cmark & \xmark &  $\mathbf{47.88}$ & $\mathbf{49.29}$ & $\mathbf{49.90}$ & $\mathbf{45.90}$ & $\mathbf{43.94}$\\
            \xmark & \xmark &  $46.12$ & $48.74$ & $48.11$ & $43.59$ & $40.99$\\
            \hline
            \cmark & $256$ & $45.22$ & $46.38$ & $46.69$ & $44.13$ & $41.83$ \\
            \cmark & $512$ & $45.09$ & $45.90$ & $47.32$ & $42.65$ & $41.98$ \\
            \cmark & $1024$ & $44.39$ & $44.98$ & $46.63$ & $41.69$ & $41.78$ \\
    \bottomrule
  \end{tabular}
  }
  \egroup
  \caption{\textbf{The Significance of Pages as Semantic Units}. Results on DUDE validation set ablating over (a) utilization of page-embedding, (b) segment length for fixed-size encoding inputs.}
  \label{table:page_ablation}
\end{table}

\begin{figure}[t]
    \centering
    \includegraphics[width=0.9\linewidth]{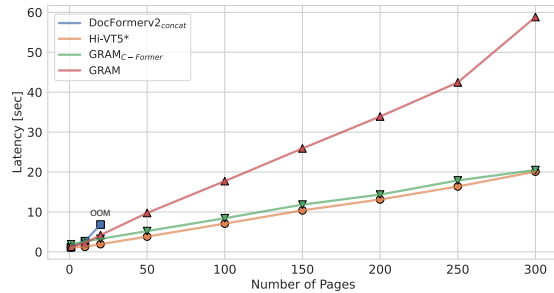}
    \caption{\textbf{Latency comparison}. We compare the dependency between overall latency and the number of pages in input document for GRAM, GRAM$_{C-Former}$, $\text{DocFormerv2}_{concat}$ and $\text{Hi-VT5}$.}
    \label{Fig:eval_latency}
\end{figure}

\paragraph{Performance-Latency Trade-off}
We assess the impact of \CFormer on performance, considering compression output lengths of ${8, 32, 256, 1024, 4096}$. Note, performance gradually improves with an increase in the compression output length. However, longer output lengths correspond to heightened model latency. 
Note that using \CFormer for shorter documents can be redundant, as there is little to no compression compared to the input sequence length and results decrease.
In \cref{Fig:eval_latency}, we scrutinize the trade-off between computational efficiency and compression rate by comparing to DocFormerv2$_{concat}$ \cite{appalaraju2023docformerv2} and Hi-VT5* \cite{tito2022hierarchical}. We discover that DocFormerv2$_{concat}$ reaches a memory limit at approximately $20$ pages, due to its quadratic memory increase with sequence length. At this juncture, GRAM$_{C-Former}$ surpasses DocFormerv2$_{concat}$ by performing $3.5$ seconds faster. Notably, GRAM$_{C-Former}$ can gracefully handle documents surpassing $300$ pages, effectively bridging the gap between performance and latency.

\section{Conclusions}
\label{sec:conclusions}

Our method, termed GRAM, extends existing single-page document models to efficiently handle multi-page documents without necessitating computationally-intensive pretraining. Leveraging the single-page encoder for local page-level comprehension, we introduce document learnable tokens and designated layers, enabling seamless information exchange across pages. Additionally, our proposed bias adaptation method enforces effective utilization of our newly introduced document tokens. The incorporation of a \CFormer model reduces sequence length, balancing quality with latency in the decoding step. Extensive experiments demonstrate GRAM's state-of-the-art performance across multi-page DocVQA benchmarks.

{
    \small
    \bibliographystyle{ieeenat_fullname}
    \bibliography{egbib}
}

\clearpage
\setcounter{page}{1}
\maketitlesupplementary

\appendix

\section{Parameters}
We present in \cref{table:parameters} all of the relevant hyperparameters.

\tabcolsep=0.11cm
\begin{table}[]
  \centering
  \bgroup
  \def\arraystretch{1.1}
  \resizebox{1\linewidth}{!}{

  \begin{tabular}{l | c  c c}
    \toprule
    \multirow{1}{*}{\textbf{Group}} & \multicolumn{1}{c}{\textbf{Parameter Name}} & \multicolumn{1}{c}{\textbf{Parameter Value}} \\
    \hline

    \multirow{7}{*}{fine-tune} & batch size  & 8 \\
     & training steps & 200K \\
     & warmup steps & 1000 \\
     & fp16 & True \\
     & training number of pages & 4 \\
     & evaluation number of pages & unlimited \\
     & number of image tokens & 128 \\

    \hline

    \multirow{4}{*}{DocFormer$_{concat}$ \cite{appalaraju2023docformerv2}} & encoder learning rate  & 3e-5 \\
     & decoder learning rate  & 3e-5 \\
     & training text tokens per page & 600 \\
     & inference text tokens per page & 400 \\

    \hline

    \multirow{5}{*}{HiVT5* \cite{tito2022hierarchical}} & encoder learning rate  & 3e-5 \\
     & decoder learning rate  & 3e-5 \\
     & training text tokens per page & 800 \\
     & inference text tokens per page & 8000 \\
     & number of compression tokens per page & 10 \\

    \hline

    \multirow{7}{*}{GRAM} & encoder learning rate  & 3e-5 \\
     & decoder learning rate  & 3e-5 \\
     & global encoder learning rate  & 1e-4 \\
     & training text tokens per page & 800 \\
     & inference text tokens per page & 8000 \\
     & number of global tokens  & 32 \\
     & bias adaptation constant `$c$' & 20 \\

    \hline

    \multirow{9}{*}{GRAM$_{C-Former}$} & encoder learning rate  & 3e-5 \\
     & decoder learning rate  & 3e-5 \\
     & global encoder learning rate  & 1e-4 \\
     & C-Former learning rate  & 1e-4 \\
     & training text tokens per page & 800 \\
     & inference text tokens per page & 8000 \\
     & number of global tokens  & 32 \\
     & bias adaptation constant `$c$' & 20 \\
     & compression length  & 256 \\
     
    \hline

    \bottomrule
  \end{tabular}
  }
  \egroup
  \caption{\textbf{Hyper-Parameters}.}
  \label{table:parameters}
\end{table}

\section{Inference Resources Consumption}
\begin{figure}[h!]
    \centering
    \includegraphics[width=0.9\linewidth]{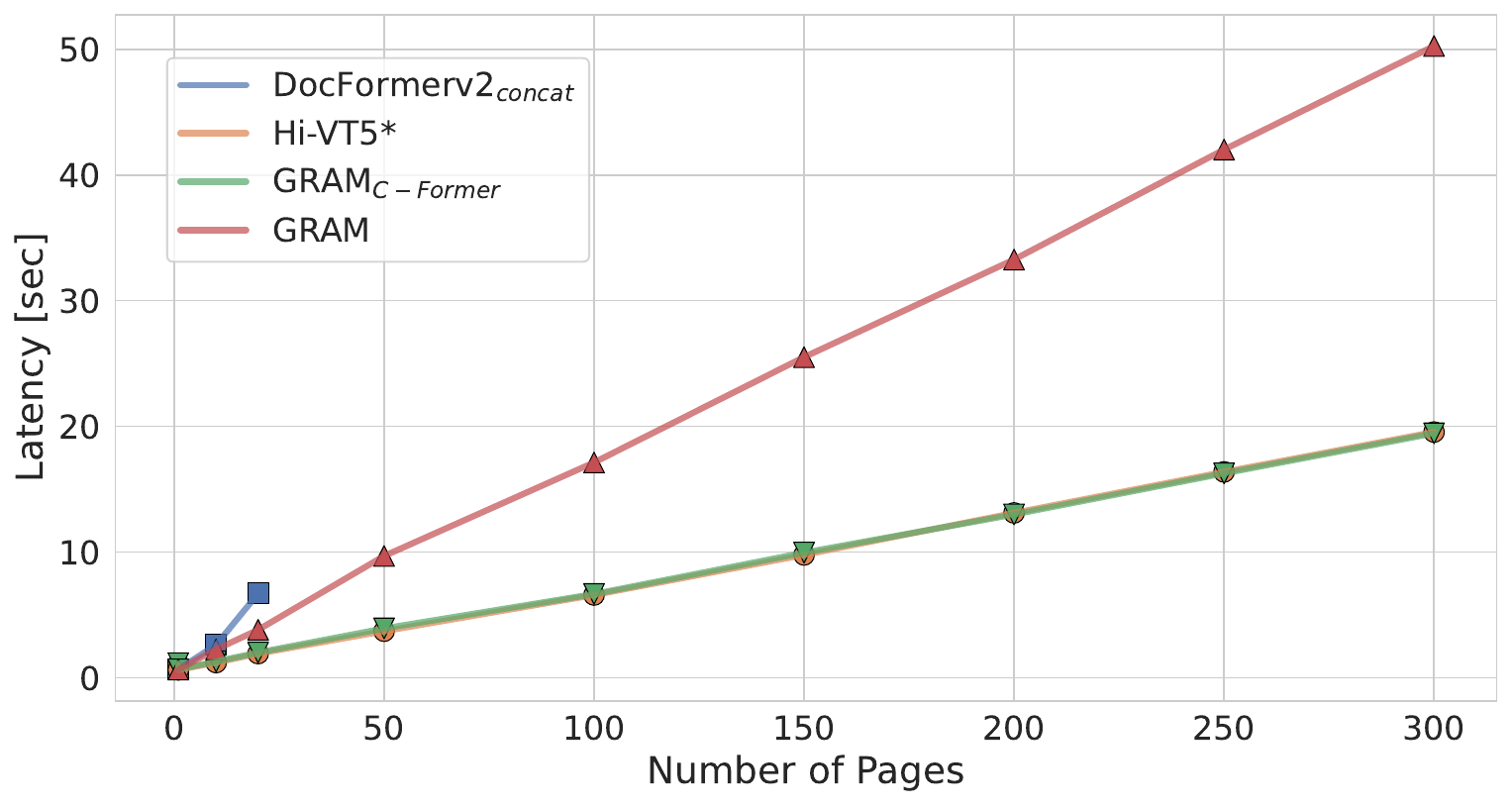}
    \caption{\textbf{Latency comparison}. We compare the dependency between overall latency and the number of pages in input document.}
    \label{Fig:app_inference_latency}
\end{figure}

\begin{figure}[h!]
    \centering
    \includegraphics[width=0.9\linewidth]{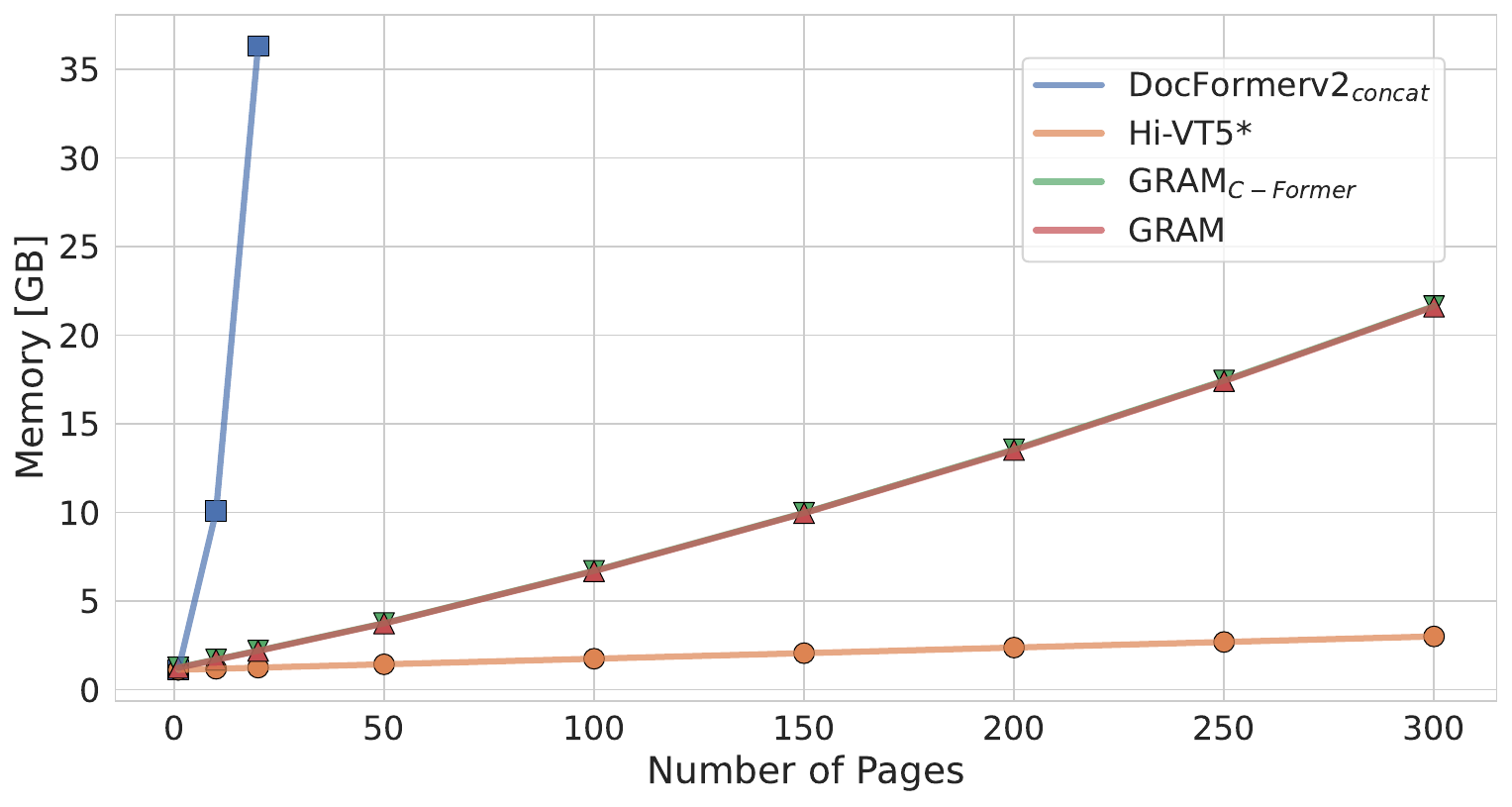}
    \caption{\textbf{Memory consumption comparison}. We compare the dependency between overall memory consumption and the number of pages in input document.}
    \label{Fig:app_inference_mem}
\end{figure}

We compare three key properties of MP-DocVQA baselines and our method: inference time, memory consumption, and maximal document length. The latency and memory consumption are illustrated in \cref{Fig:app_inference_latency} and \cref{Fig:app_inference_mem}, respectively, both as functions of the number of pages in the document. We compare the following baselines: DocFormerv2$_{concat}$ \cite{appalaraju2023docformerv2}, \Hivt* \cite{tito2022hierarchical}, and our GRAM and GRAM$_{C-Former}$, utilizing the same computational resources employed in all experiments— $8\times A100$ GPUs with $40GB$ of memory.

The memory consumption of DocFormerv2$_{concat}$ \cite{appalaraju2023docformerv2} reaches its maximum capacity for documents with only $20$ pages, while our method efficiently processes documents, spanning hundreds of pages. Moreover, the presented figures demonstrate that GRAM$_{C-Former}$ maintains a comparable memory footprint to the GRAM model. Nevertheless, there is potential for improvement, as HiVT5* exhibits lower memory consumption. Despite this, we achieve inference times similar to HiVT5* \cite{tito2022hierarchical}, accompanied by a noteworthy enhancement in ANLS.

\section{Qualitative Results}
\begin{figure*}[ht]
    \centering
    \includegraphics[width=0.9\textwidth]{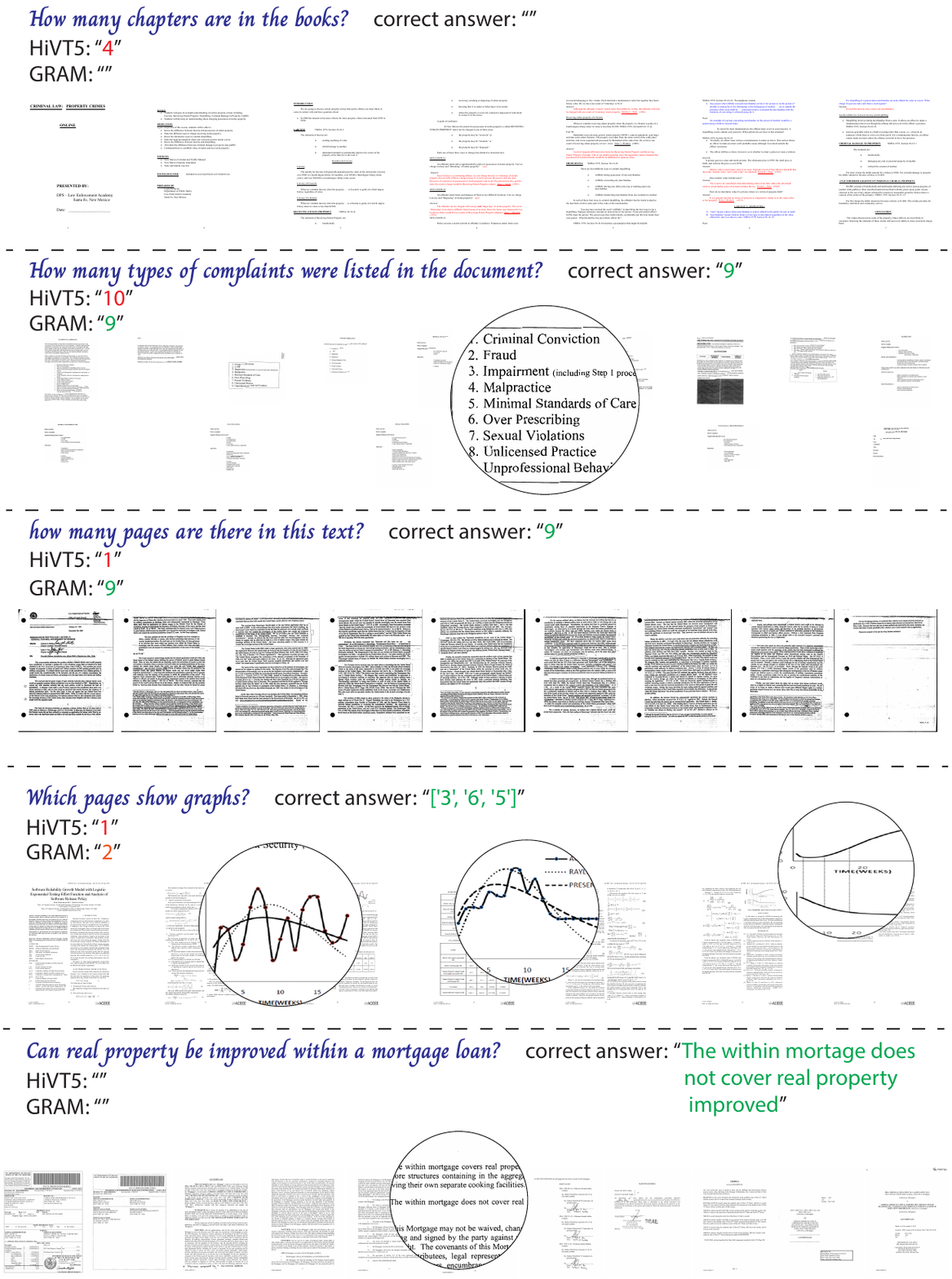}
    \caption{Qualitative comparison between our approach and \Hivt \cite{tito2022hierarchical} indicates that the integration of our global-local encoder enhances reasoning capabilities, especially when inquiries require multi-page context.}

    \label{Fig:app_quantitative_results}
\end{figure*}

Finally, we present a few qualitative results on the DUDE dataset in \cref{Fig:app_quantitative_results}, showcasing the advantages of our approach over Hi-VT5* \cite{tito2022hierarchical}. In the first three examples, we demonstrate cases where GRAM is correct and HiVT5* is wrong. The last two examples present cases where both our method and HiVT5* are incorrect.

\section{Comparison with $\text{DocFormerV2}_{\text{concat}}$}

We provide additional qualitative examples with $\text{DocFormerV2}_{\text{concat}}$. Examples demonstrate the effectiveness of GRAM in tackling questions that involve multiple pages in the document.

\begin{figure*}[]
    \centering
    \includegraphics[width=1.0\linewidth]{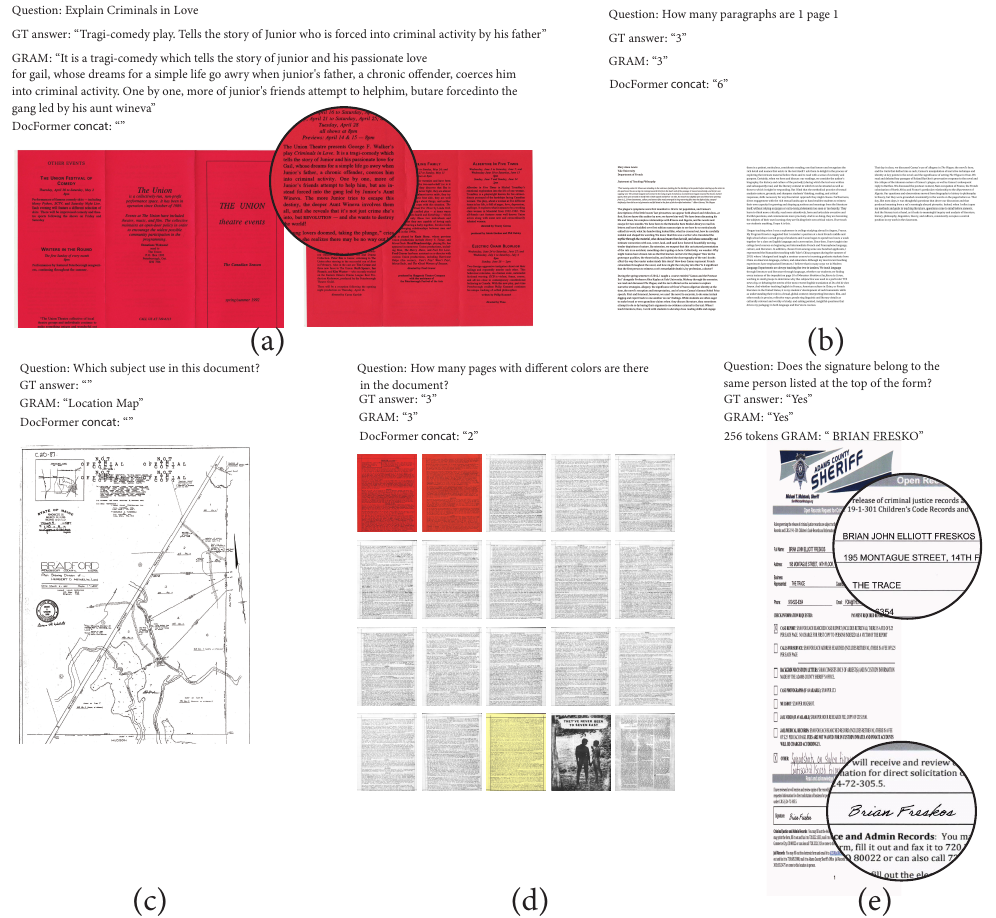}
    \caption{Comparisons between $\text{DocFormerV2}_{\text{concat}}$ and GRAM.}
    \label{Fig:examples}
\end{figure*}

\section{Comparison with NLP-based Approaches}
We present additional experiments, comparing GRAM with two NLP-based approaches: the sparse attention-based LongFormer \cite{beltagy2020longformer}, and the bias-based  AliBi \cite{press2021train}. Both approaches are implemented on top of DocFormerv2 for fair comparison. Results in \cref{table:nlp_methods} shows an advantage in our local-global approach of utilizing existing powerful models for single-page and extending them to support the multi-page scenario.

\tabcolsep=0.09cm
\begin{table}[]
  \centering
  \bgroup
  \def\arraystretch{1.0}
  \resizebox{1\linewidth}{!}{
  \begin{tabular}{l | c c c c c}
    \toprule
        \multirow{3}{*}{\textbf{\shortstack{Method}}} & 
        \multicolumn{5}{c}{\textbf{ANLS by Number of Pages}} \\
        & 
        \multicolumn{5}{c}{\textbf{DUDE validation dataset}} \\
        & All & 1 & 2-4 & 5-10 & 11-end \\
            \hline
            \rowcolor{golden!20}
            GRAM &  $\mathbf{47.88}$ & $\mathbf{49.29}$ & $\mathbf{49.90}$ & $\mathbf{45.90}$ & $\mathbf{43.94}$\\
            $\text{DocFormerv2}_{\text{concat}}$ &  $44.32$ & $46.08$ & $47.05$ & $42.81$ & $38.17$ \\
            $\text{DocFormerV2}_{\text{Longformer}}$ &  $45.88$ & $47.01$ & $47.75$ & $43.22$ & $43.13$ \\
            $\text{DocFormerV2}_{\text{AliBi}}$ &  $34.73$ & $36.55$ & $37.00$ & $30.99$ & $31.25$ \\
    \bottomrule
  \end{tabular}
  }
  \egroup
  \caption{\textbf{Comparison to NLP methods}. Results on DUDE validation comparing GRAM with LongFormer \cite{beltagy2020longformer} and AliBi \cite{press2021train}.}
  \label{table:nlp_methods}
\end{table}


\end{document}